\documentclass[conference]{IEEEtran}

\usepackage{color}
\usepackage[ruled,linesnumbered,lined]{algorithm2e}
\usepackage{graphicx}
\usepackage{subcaption}
\date{}
\input{Qcircuit}

\newcommand{\tens}{%
 $ \mathbin{\mathop{\otimes}\displaylimits}$%
}

\begin{document}
\title{Neural Networks Architecture Evaluation in a Quantum Computer}

\author{\IEEEauthorblockN{Adenilton J. da Silva, and Rodolfo Luan F. de Oliveira}
\IEEEauthorblockA{Departamento de Estat\'{i}tica e Inform\'{a}tica\\
Universidade Federal Rural de Pernambuco\\
Recife, Pernambuco, Brazil 52171-900 \\
 Email: \{adenilton.silva, rodolfo.luanoliveira\}@ufrpe.br}}

\maketitle

\begin{abstract}
In this work, we propose a quantum algorithm to evaluate neural networks architectures named Quantum Neural Network Architecture Evaluation (QNNAE). 
The proposed algorithm is based on a quantum associative memory and the learning algorithm for artificial neural networks. 
Unlike conventional algorithms for evaluating neural network architectures, QNNAE does not depend on initialization of weights. 
The proposed algorithm has a binary output and results in 0 with  probability proportional to the performance of the network. 
And its computational cost is equal to the computational cost to train a neural network.
\end{abstract}

\IEEEpeerreviewmaketitle

\section{Introduction}

There are some problems that quantum computing provides time complexity advantages over classical algorithms. 
For instance, Grover's search algorithm~\cite{grover:96} has a quadratic gain when compared with the best classical algorithm and the Shor's factoring algorithm~\cite{shor:97} provides an exponential gain when compared with the best known classical algorithm. 
Quantum computing also provides space complexity advantages over classical methods. 
For instance, a quantum associative memory has an exponential gain in storage capacity when compared with classical associative memories~\cite{ventura2000quantum,trugenberger2001probabilistic}.

Artificial Neural Networks~\cite{haykin:99} (ANN) are another model of computation that provides advantages over classical algorithms.
ANN have the ability to learn from the environment and can be used to solve some problems that do not have known algorithmic solution.

The development of machine learning applications requires an empirical search over  parameters. 
Several techniques have been proposed to perform parameter selection. In~\cite{miranda2014hybrid} a meta-learning strategy is used to suggest initial SVM configurations.  
Evolutionary computation~\cite{pacifico2012improved} and several other metaheuristics~\cite{ojha2017metaheuristic}  have been used to select artificial neural networks.
In~\cite{daSilva2016quantum} quantum computation combined with a nonlinear quantum operator is used to select a neural network architecture.

In this work, we suppose that a quantum computer with thousands of qubits will be created and it will follow the rules of quantum computation described in~\cite{nielsen:00}.  
In previous works, nonlinear quantum computation~\cite{PhysRevLett.81.3992} has been used in the learning algorithm of a neural network~\cite{panella:09, daSilva2016quantum}.
We do not know if this supposition is realistic and we avoid any nonlinear or non-unitary operators.

It is possible to use quantum computers to obtain a global information about a function with only one function call~\cite{deutsch1992rapid}. The objective of this work is to propose a quantum algorithm to evaluate a global information about neural networks architectures. The quantum algorithm evaluates a neural network architecture training the neural network only once.
The proposed algorithm has binary output and results in 0 with high probability if the neural network can learn the data set with high performance.
The computational complexity of the algorithm is equal to the computational complexity of a neural network learning algorithm.
The algorithm is based on the quantum superposition based learning strategy~\cite{ventura:04,panella:09} and on the quantum associative proposed by Trugenberg~\cite{Trugenberger2002}.

The remainder of this paper is divided into 5 sections. 
Section~\ref{sec:qc} presents the basic concepts of quantum computing necessary to the development of this work. 
Section~\ref{sec:qam} presents quantum associative memory used in the proposed algorithm. 
Section~\ref{sec:la} presents the main contribution of this work: a quantum algorithm to evaluate artificial neural networks architectures that does not depends on weights initialization.
Section~\ref{sec:exp} presents an empirical evaluation of the proposed algorithm. This evaluation was performed using the classical version of the algorithm. Section~\ref{sec:conclusion} finally presents the conclusion and future works.

\section{Quantum computation}
\label{sec:qc}
The quantum bit (qubit) is a two-level quantum system that has unitary evolution  over time in a 2-dimensional complex vector space. 
An orthonormal basis (named computational basis) in this space is described in Eq.~\eqref{eq:kt}. 
One qubit can be described as a superposition, linear combination,  of the vectors in the computational basis, as described in Eq.~\eqref{eq:sup}, in which $\alpha$ and $\beta$ are complex amplitudes conditioned with the following normalization $|\alpha|^2 + |\beta|^2 = 1$. 
This feature causes a quantum bit to be represented in many ways, unlike a classical bit that is always represented by 0 or 1. 

\begin{equation}
\ket{0} =  \begin{bmatrix}
1 \\
0
\end{bmatrix}
\ket{1} =  \begin{bmatrix}
0 \\
1
\end{bmatrix}
\label{eq:kt}
\end{equation}

\begin{equation}
\ket{\psi}=\alpha \ket{0} + \beta\ket{1}
\label{eq:sup}
\end{equation}

For a system with several qubits the tensorial product (\tens) is used. 
Given two quantum bit $\ket{\psi} =\alpha \ket{0} + \beta \ket{1}$   and $\ket{\phi} =\rho \ket{0} + \theta\ket{1}$  the tensor product $\ket{\psi}$ \tens $\ket{\phi}$  is equal to the vector $\ket{\psi\phi}$ = $\alpha\rho\ket{00} + \alpha \theta\ket{01} + \rho \beta\ket{10} + \beta \theta\ket{11}$. 

\subsection{Operators}
A quantum operator on $n$ qubits is represented by a $2^n \times  2^n$ unitary matrix. 
An $M$ matrix is  unitary if  $MM^\dag = M^\dag M = I$, where $ M^\dag$ is the conjugate transpose of $M$ and $I$ is the identity matrix. 
Any $2^n \times  2^n$ unitary  matrix describes a valid quantum operator on $n$ qubits. 
An example of a quantum operator, named Hadamard, and its actions on the computational basis are shown in Eq.~\eqref{eq:had}.

\begin{equation}
\label{eq:had}
\begin{minipage}{0.15\textwidth}
$
H = \frac{1}{\sqrt{2}}\begin{bmatrix}
1 & 1 \\
1 & -1
\end{bmatrix}
$
\end{minipage}
\begin{minipage}{0.25\textwidth}
$
H\ket{0} = \frac{1}{\sqrt{2}} \left( \ket{0} + \ket{1}\right)
$

$H\ket{1} = \frac{1}{\sqrt{2}} \left( \ket{0} - \ket{1}\right)$
\end{minipage}
\end{equation}

Not surprisingly, quantum operators can simulate classical operators. 
The key to this is the Toffoli operator described in Eq.~\eqref{eq:tof}. 
Every classical operator can be decomposed into Nand operators, which is irreversible. 
However, even though the Nand operator is irreversible, with proper treatment, the Toffoli operator is able to simulate it.
So quantum operators can simulate any binary function whether it is reversible or not.

\begin{equation}
T\ket{x,y,z} = \ket{x,y,z \oplus (x\cdot y)}
\label{eq:tof}
\end{equation}

\subsection{Quantum parallelism}

The ability of intrinsic parallelism is one of the most promising characteristics of quantum computation. 
For instance, given a function $f(x): \{0,1\} \rightarrow \{0,1\}$,  it is possible to create  a unitary operator $U_f$ that implements $f$ as described in Eq.~\eqref{eq:pare}. 
It is possible to verify several inputs as described in Eq.~\eqref{eq:paresup}, where $x = H\ket{0}$ and $y = \ket{0}$.

\begin{equation}
U_f\ket{x,y}= \ket{x,y \oplus f(x)}
\label{eq:pare}
\end{equation}

\begin{equation}
U_f\left(\frac{\ket{0, 0}+ \ket{1, 0}}{\sqrt{2}}\right) = \frac{\ket{0, f(0)}+ \ket{1, f(1)}}{\sqrt{2}}
\label{eq:paresup}
\end{equation}

\subsection{Measuring}
Almost all quantum operators are unitary, except the measurement operation. 
Unlike classical computing, in which values can be measured and the system remains unchanged, in quantum computing, the very act of measuring causes the system to change. 
After a measurement the quantum state to collapse to one of its possible values.
For instance, if a quantum system is in the state described in Eq.~\eqref{eq2qubit}.
\begin{equation}
\alpha_{00}\ket{00} + \alpha_{01}\ket{01} + \alpha_{10}\ket{10} + \alpha_{11}\ket{11} 
\label{eq2qubit}
\end{equation}
After the measurement the state will collapse to $\ket{x_1x_2}$ with probability $\left| \alpha_{x_1x_2} \right|^2$.

\section{Quantum associative memory}
\label{sec:qam}
The quantum associative memory used in this work was proposed in \cite{trugenberger2001probabilistic}. 
This associative memory functionality can be divided into two phases: i) quantum storage mechanism and ii) quantum retrieval mechanism.
The algorithm of the storage phase used in~\cite{trugenberger2001probabilistic} is equivalent to the algorithm proposed in \cite{ventura2000quantum}  that presents the first model of quantum associative  memory. Given a dataset with $p$ patterns $\{m^{p_k}\}_{{p_k}=1}^p$ the storage algorithm creates the quantum state described in Eq.~\eqref{eq:memory}.
\begin{equation}
\sum_{p_k = 1}^p  \frac{1}{\sqrt{p}}\ket{m^{p_k}}.
\label{eq:memory}
\end{equation}

In the retrieval phase, the quantum memory receives  an input pattern, which may be a corrupted version of a pattern already stored and probabilistically indicates the chance of the memory containing the pattern. 
The main characteristic of this phase is that it does not require a classic auxiliary memory, as is the case of Ventura's memory.
This probability of recognition of the presented pattern is accessed by measuring the control qubit $\ket{c}$ observing the probability of being $\ket{0}$.

\begin{equation}
\ket{\psi} = \frac{1}{\sqrt{p}}\sum_{p_k=1}^p \ket{i;m^{p_k};c}
\label{eq:sta}
\end{equation}

As described  in Eq.~\eqref{eq:sta} the state in the recovery phase can be divided into three parts. 
The quantum register $i$, of size $n$, represents the input to be checked. The value $m^{p_k}$ representing the $p$ values stored in the storage phase and the quantum register $c$, initialized with $H\ket{0}$, is the control qubit. 

After the execution of  probabilistic quantum memory retrieval algorithm, measuring the control qubit $\ket{c}$ will result $\ket{c} = \ket{0}$ with probability described in Eq.~\eqref{eq:p0}. 
The retrieval algorithm of Trugenberg's probabilistic memory is described in~\cite{trugenberger2001probabilistic}.

\begin{equation}
P\left(\ket{c} = \ket{0}\right) = \frac{1}{p}\sum_{k=1}^p cos^2\left(\frac{\pi}{2n}d_H(i,p^k)\right)
\label{eq:p0}
\end{equation}

\begin{equation}
P\left(\ket{c} = \ket{1}\right) = \frac{1}{p}\sum_{k=1}^p sin^2\left(\frac{\pi}{2n}d_H(i,p^k)\right)
\label{eq:p1}
\end{equation}

As described in Eq.~\eqref{eq:p0} and Eq.~\eqref{eq:p1} the probability of recognition, or not, is related to the proximity of the input to the stored patterns. 
The measure of similarity used is the Hamming distance represented by $d_H(i,p^k)$. 
When looking for an isolated pattern, the distance between the patterns will be, for the most part, greater than 0, collaborating for ket $\ket{c} = \ket{1}$.

\section{Quantum learning algorithm}
\label{sec:la}

Quantum neural networks~\cite{schuld2015simulating, panella2009neurofuzzy,schuld2014quest,NIPS2003_2363,NIPS2016_6401,wiebe2014quantum,altaisky2014quantum} and quantum inspired neural networks~\cite{gao2017deep, li2013hybrid,cardoso2015quantum} have been proposed in several works. In this work, we follow a different approach and analyze the possibility to obtain global properties of a classical neural network using a quantum algorithm.

Evaluate a neural architecture is not an easy task, ``the mapping from an architecture to its performance is indirect, ..., and dependent on the evaluation method used'' \cite{yao1999evolving}. The
number of hidden layers, neurons per layers, the activation function is determined by the experience of the researcher~\cite{ding2013evolutionary} and there is no algorithm to determine the optimal neural network architecture~\cite{abraham2004meta}.

In this Section, we describe the proposed quantum algorithm that uses quantum parallelism to evaluate neural networks architectures (number of neurons in hidden layer). 
Several works  propose techniques to search for near-optimal neural networks architectures~\cite{yao1999evolving,ding2013evolutionary,abraham2004meta,ojha2017metaheuristic}. One limitation of these works is the impossibility to evaluate a neural network architecture without many random weights initialization~\cite{ding2013evolutionary}. 
The quantum algorithm proposed in this section performs a neural network architecture evaluation that does not depends on a weights initialization because quantum parallelism is used to initialize the neural network weights with all possible weights.

Classically the main idea of this algorithm is to train all possible neural networks with a given architecture and create a binary vector $performance_j$ for each weight vector $\textbf{w}_j$. 
The $perfomance_j$ has the same size of a validation dataset $\left\{ \left(x_k, d_k\right)\right\}$ and the position $k$ of the vector is equal to 1 if the network with weights $\textbf{w}_j$ correctly classify $x_k$.
The algorithm compares the performance of each neural network with the performance  representing an accuracy of 100\% as described in Eq.~\eqref{eq:p}, where $ts$ is the number of qubits in the validation set, $|W|$ is the number of weights in superposition and $d_H\left(\cdot\right)$ calculates the hamming distance.
\begin{equation}
\sum_{k=1}^{|W|} \frac{1}{|W|} \cos^2 \left( \frac{\pi}{2t_s} \cdot d_H(\ket{1}_{ts}, \ket{performance_k}) \right)
\label{eq:p}
\end{equation}
In a classical computer, this strategy cannot be accomplished in polynomial time. 
We show that in a quantum computer this algorithm can be executed in polynomial time and the result is related to the neural network architecture capacity to learn a dataset. 

The quantum version of the proposed algorithm consists in create a superposition of all neural networks. Train the neural networks in the superposition. Create a performance vector for each neural network in the superposition and use the recovering algorithm of the quantum associative memory using the performance quantum register as the quantum memory and with input $\ket{1}_{ts}$ to evaluate the neural network architecture. 
This idea is precisely described in Algorithm~\ref{alg:ajsalgorithm}.

\begin{algorithm}
Initialize all weights qubits with $H\ket{0}$  \label{line:1}\\ 
Divide the dataset $\mathcal{T}$ in a train set and a validation dataset \label{line:2}\\
Initialize quantum register $\ket{performance}$ with the quantum register $\ket{0}_n$ \label{line:3}\\
Train the neural networks in superposition with the train set \label{line:4}\\
\For{\textbf{each} pattern $p_j$ in and desired output $d_j$ in the validation set\label{line:5}}{
Initialize the quantum registers $p,o,d$ with the quantum state $\ket{p,0,d}$ \label{line:6}\\
Calculate $N\ket{p_k}$ to calculate network output in quantum register $\ket{o}$ \label{line:7}\\
\If{$\ket{o} = \ket{d}$ \label{line:8}}
	{Set $\ket{performance}_j$ to 1 \label{line:9}}
Calculate $N^{-1}$ to restore $\ket{o}$ \label{line:10}
}\label{line:11}
Apply the quantum associative recovering algorithm with input $\ket{1}_n$ and memory $\ket{performance}$ \label{line:13}\\
Return the control qubit $\ket{c}$ of the quantum associative memory
\caption{Evaluate architecture}
\label{alg:ajsalgorithm}
\label{alg:grover}
\end{algorithm}

Lines~\ref{line:1} to~\ref{line:3} of Algorithm~\ref{alg:ajsalgorithm} performs the initialization. 
Line~\ref{line:1} creates a superposition of all possible neural networks for a given architecture.
Line~\ref{line:2} splits a dataset $\mathcal{T}$ in a train set and a test set. 
We suppose the $\mathcal{T}$ is composed of classical data (or data in computational basis). 
Line~\ref{line:3} initialise quantum register performance with the state $\ket{0}_{ts}$, where $ts$ is the number of patterns in the validation set.

Line~\ref{line:4} trains the neural networks in the superposition. 
The learning algorithm of a classical neural network can be viewed as a binary function that sends the input $x(t)$, desired output $d(t)$ and weights $w(t)$ in iteration $t$ to the new weight vector $w(t+1)$ as described in Eq.~\eqref{eq:weightupdate}.
\begin{equation}
w(t+1) = f\left(w\left(t\right),x\left(t\right), d\left(t\right)\right)
\label{eq:weightupdate}
\end{equation}
Any binary function can be simulated by a quantum operator $U_f$ then the learning procedure can be simulated in a quantum computer and it is possible to train several  artificial neural networks using quantum parallelism.

The for loop in line \ref{line:5} to line \ref{line:11} evaluates the performance of each neural network in the superposition. After this loop, each neural network is entangled with its performance vector that has 1 in position $j$ if and only if the network correctly classifies the $j$th vector in the validation dataset.

In line~\ref{line:13} the recovering algorithm of the probabilistic quantum memory is applied to the performance quantum register as memory and input $\ket{1}_{ts}$. The input $\ket{1}_{ts}$ represents a performance of 100\%. Line 14 returns the control bit $\ket{c}$ of the probabilistic quantum memory. A measurement of $\ket{c}$ will return 0 with high probability if the neural network architecture is capable of learning the dataset.

\begin{figure*}[!h]
\centering
\begin{subfigure}{.49\textwidth}
  \centering
  \includegraphics[width=1.1\linewidth]{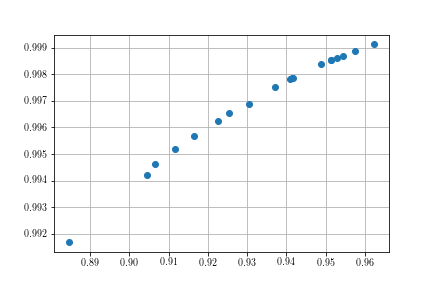}
  \caption{}
  \label{fig:sub11}
\end{subfigure}%
\begin{subfigure}{.49\textwidth}
  \centering
  \includegraphics[width=1.1\linewidth]{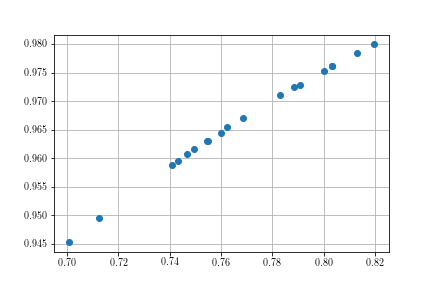}
  \caption{}
  \label{fig:sub12}
\end{subfigure}
\begin{subfigure}{.49\textwidth}
  \centering
  \includegraphics[width=1.1\linewidth]{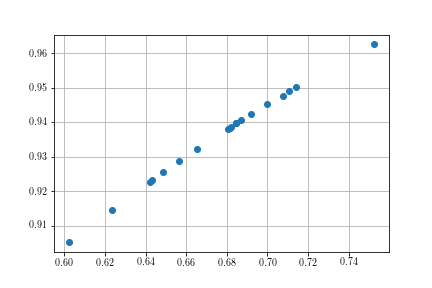}
  \caption{}
  \label{fig:sub13}
\end{subfigure}
\begin{subfigure}{.49\textwidth}
  \centering
  \includegraphics[width=1.1\linewidth]{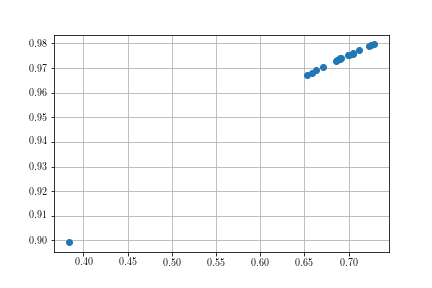}
  \caption{}
  \label{fig:sub14}
  \end{subfigure}
\caption{Artificial neural network performance versus output of QNNAE algorithm $P\left(\ket{c}\right) = \ket{0}$ in datasets (a) cancer, (b) card, (c) diabetes and (d) gene.}
\label{fig:1}
\end{figure*}

\begin{figure*}[h]
\centering
  \begin{subfigure}{.49\textwidth}
  \centering
  \includegraphics[width=1.1\linewidth]{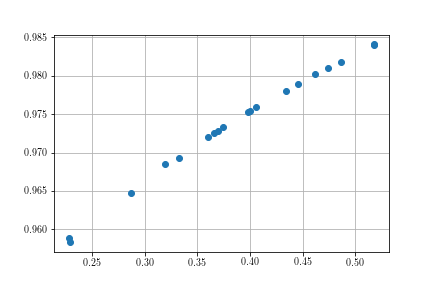}
  \caption{}
  \label{fig:sub1}
\end{subfigure}
  \begin{subfigure}{.49\textwidth}
  \centering
  \includegraphics[width=1.1\linewidth]{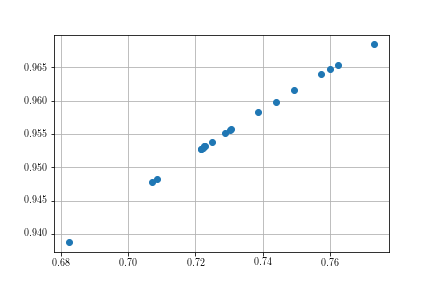}
  \caption{}
  \label{fig:sub2}
\end{subfigure}
  \begin{subfigure}{.49\textwidth}
  \centering
  \includegraphics[width=1.1\linewidth]{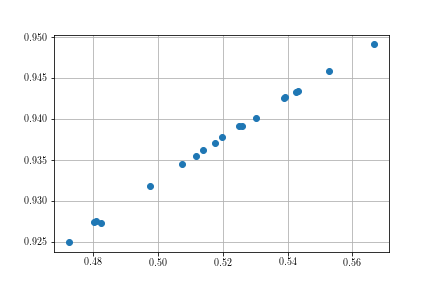}
  \caption{}
  \label{fig:sub3}
\end{subfigure}
  \begin{subfigure}{.49\textwidth}
  \centering
  \includegraphics[width=1.1\linewidth]{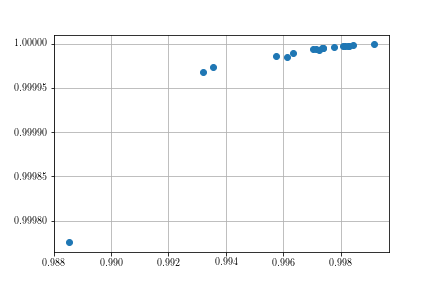}
  \caption{}
  \label{fig:sub4}
\end{subfigure}
\caption{Artificial neural network performance versus output of QNNAE algorithm $P\left(\ket{c}\right) = \ket{0}$ in datasets (a) glass, (b) heart, (c) horse and (d) mushroom.}
\label{fig:2}

\end{figure*}

The computational cost of Algorithm 2 is equal to the cost to train a single neural network and to evaluate its performance in the validation dataset. 
Lines 1 to 3 have computational cost $O(n_w + p)$, where $n_w$ is the number of weights and $p$ is the number of patterns in $\mathcal{T}$.
Line 4 trains a neural network and has the same cost of the neural networks learning procedure.
The for loop in lines 5 to 12 evaluate the performance of neural networks in the superposition with cost $O(t_s)$. Line 13 applies the quantum associative memory with cost $O(n_w)$ and in line 14 the measurement has cost O(1). 

\section{Experiments}
\label{sec:exp}
A quantum computer necessary to perform an experimental evaluation of this work is not yet a reality. 
Simulation of Algorithm 2 with actual technology is impossible  because the simulation of quantum systems in classical computers has exponential cost in relation to the number of qubits.
For instance, in~\cite{haner20170} the simulation of 45 qubits requires 0.5 petabits of memory.

The neural network simulation and algorithm output were simulated on a classical computer. 
To perform an evaluation of Algorithm 2 we follow its classical description and performs modifications of the number of neural networks in the quantum parallelism.
This modification was necessary due to the high cost of train neural networks with all possible weights initialization.
Instead of using all possible weights, we randomly select 1000 neural network weight vectors for a single layer neural network with $k$ hidden neurons.

To evaluate the proposed algorithm we used the datasets described in Section~\ref{sec:datasets}.
We split each dataset into train and validation sets.
The train sets contain 10\% of the patterns in the dataset and the validation dataset contains 90\% of the patterns.
The train sets were used to train the neural networks and the validation sets were used to compute the performance vectors.

For each dataset and for each number of hidden neurons in the interval $\left[1,20\right)$ 1000 single layer neural networks were trained  and the performance vectors were calculated using the validation dataset. 
The simulation was performed with the multi-layer neural network available in scikit-learn~\cite{scikit-learn} version 0.18. 
The neural networks parameters used in the experiment are displayed in Table~\ref{table:param}.
The probability of the algorithm output 0 was calculated using Eq.~\eqref{eq:p0}, that represents the output of a  quantum associative memory with input $\ket{1}_{ts}$ and quantum memory consisting of the  network performances in superposition.

\begin{table}[!t]
\caption{Neural network learning algorithm parameters}
\label{table:param}
\centering
\begin{tabular}{|c||c|}
\hline
Parameter & value  \\ 
\hline
learning algorithm & lbfgs\\
\hline
alpha & 1e-5\\
\hline
number of hidden neurons & $\left[1, 20\right]$\\
\hline
max{\_}iter & 400\\
\hline
\end{tabular}
\end{table}


\subsection{Datasets}
\label{sec:datasets}
To evaluate QNNAE, we perform experiments with eight well-known classification problems found in PROBEN1 repository. 
The datasets are: cancer, card, diabetes, gene, glass, heart, horse and mushroom. 
Table~\ref{table:data} describe some characteristics of these datasets.

\begin{table}[!b]
\caption{Datasets characteristics}
\label{table:data}
\centering
\begin{tabular}{|c|c|c|c|}
\hline
Problem & examples & features & classes  \\ 
\hline
Cancer & 699 & 9 & 2\\
\hline
Card & 690 & 51 & 2\\
\hline
Diabetes & 768 & 8 & 2\\
\hline
Gene & 3175 & 120 & 3\\ \hline
Glass & 214 & 9 & 6\\ \hline
Heart & 920 & 35 & 2 \\ \hline
Horse & 364 & 58 & 3\\ \hline
Mushroom & 8124 & 125 & 2\\ \hline

\end{tabular}
\end{table}

\subsection{Results}

Results are displayed in Fig.~\ref{fig:1} and Fig.~\ref{fig:2}, where the horizontal axis represents neural network architecture average performance and the vertical axis means the probability of the proposed algorithm outputs 0.
Each point represents the  architecture of a neural network.
It is evident the relation between the output of Algorithm 1 and neural network performance and the probability of the output 0 gives a measurement of the neural network architecture performance.

These results suggest that the binary output of QNNAE is directly related to the mean performance of the neural network architecture initialized with different weights. 
To evaluate the neural network architecture is necessary to repeat the algorithm  $\kappa$ times to estimate $P(\ket{c} = \ket{0})$.

\section{Conclusion}
\label{sec:conclusion}
We proposed a quantum algorithm that evaluates a neural network architecture performance over a dataset. 
The  proposed algorithm is the first algorithm to perform such task using quantum computation. 
The evaluation of the neural network does not depend on a weight initialization because quantum parallelism is used to create a superposition with all possible weights.
This kind of evaluation is intractable in a classical computer.

There is a lot of space for further research. 
There is a relation between the neural network performance and the probability of Algorithm 2 output 0, but the range of probabilities can be very close.
How can this methodology  be changed to improve the range of probabilities?
The choice of other neural networks parameters can be evaluated using the same framework, for instance, the choice of learning algorithm and learning algorithm parameters?
Can we evaluate the output of a quantum ensemble of classifiers using an associative memory? 
This article provides an initial step to evaluate neural networks architectures using (standard) quantum computing.
The main difference to the previous approach~\cite{daSilva2016quantum} is the use of only unitary quantum operators. 

The proposed methodology can also be applied to others machine learning methods. 
For instance, the methodology can be applied to evaluate different machine learning models over a dataset.
This strategy could lead to a quantum meta-learning approach.

\section*{Acknowledgment}
This work is supported by research grants from CNPq and FACEPE (Brazilian research agencies).

\bibliographystyle{IEEEtran}
\bibliography{bibliografia}

\end{document}